# AVP-Fusion: Adaptive Multi-Modal Fusion and Contrastive Learning for Two-Stage Antiviral Peptide Identification

Xinru Wen, Weizhong Lin, and Xuan Xiao[*]

Corresponding: author.Xuan Xiao xiaoxuan@jci.edu.cn

## Abstract

Accurate identification of antiviral peptides (AVPs) is critical for accelerating novel drug development. However, current computational methods struggle to capture intricate sequence dependencies and effectively handle ambiguous, hard-to-classify samples. To address these challenges, we propose AVP-Fusion, a novel two-stage deep learning framework integrating adaptive feature fusion and contrastive learning. Unlike traditional static feature concatenation, we construct a panoramic feature space using 10 distinct descriptors and introduce an Adaptive Gating Mechanism.This mechanism dynamically regulates the weights of local motifs extracted by CNNs and global dependencies captured by BiLSTMs based on sequence context. Furthermore, to address data distribution challenges, we employ a contrastive learning strategy driven by Online Hard Example Mining (OHEM) and BLOSUM62-based data augmentation, which significantly sharpens the model's decision boundaries. Experimental results on the benchmark Set 1 dataset demonstrate that AVP-Fusion achieves an accuracy of 0.9531 and an MCC of 0.9064, significantly outperforming state-of-the-art methods. In the second stage, leveraging transfer learning, the model enables precise subclass prediction for six viral families and eight specific viruses, even under limited sample sizes. In summary, AVP-Fusion serves as a robust and interpretable tool for high-throughput antiviral drug screening.

**Keywords:** Antiviral peptides; Two-stage prediction; Contrastive learning; Transfer learning; Biological mutation; Self-attention mechanism; Adaptive gating mechanism

## Introduction

Viral diseases and the growing challenge of drug resistance pose severe threats to global public health, highlighting the urgent need for novel therapeutics[1-5]. Antiviral peptides (AVPs) have emerged as promising alternatives to traditional drugs due to their high specificity and low toxicity[6-8]. Consequently, machine learning (ML) has been extensively applied to accelerate AVP discovery. Early research focused on traditional ML algorithms. For instance, Thakur et al. [9]pioneered AVPpred, utilizing Support Vector Machines (SVM) with amino acid composition (AAC) and physicochemical properties. Subsequently, Chang et al. [10]improved performance using Random Forest (RF) based on aggregation propensity, while Lissabet et al. introduced AntiVPP 1.0[11], integrating features like electrostatic charge and hydrophobicity. To enhance robustness, ensemble learning strategies were adopted; notably, Schaduangrat et al. developed Meta-iAVP[12], a meta-predictor aggregating six learning hypotheses, and Akbar et al. combined discrete wavelet transform with SHAP-based feature selection to construct an optimized ensemble classifier.

Despite these advancements, traditional ML approaches face significant limitations. First, reliance on outdated benchmark datasets (e.g., those from 2012) fails to reflect the current sequence diversity found in modern AVP databases. Second, heavy dependence on manual feature engineering (e.g., AAC, PseAAC) requires extensive domain expertise and often overlooks deep evolutionary information or complex molecular interactions. Third, most studies restrict themselves to binary classification, neglecting the critical need to predict specific activities against distinct viral families. Although Pang et al. attempted a two-stage model to address subclass classification, issues with generalization and efficiency remain.

To overcome these bottlenecks, deep learning (DL) has been introduced for its powerful feature extraction capabilities. Li et al. proposed DeepAVP[13], a dual-channel network combining CNN[14] and LSTM[15] to capture complex sequence patterns, while ENNAVIA[16] utilized neural networks for physicochemical feature learning. However, current DL methods often rely on static feature concatenation—ignoring dynamic context dependencies—and employ random negative sampling, which leads to blurred decision boundaries. To address these gaps, we propose AVP-Fusion, a novel two-stage framework. By synergizing an adaptive gating mechanism [17] for multimodal feature fusion with a biologically informed "second-best" mutation strategy (guided by BLOSUM62[18, 19]) and prototype-based Online Hard Example Mining (OHEM)[20], AVP-Fusion efficiently identifies AVPs and precisely predicts their functional subclasses, providing a robust tool for rational peptide design.

# Results

## Evaluation metrics

To comprehensively and objectively evaluate the performance of the AVP-Fusion model in both first-stage and second-stage tasks, we adopted a series of standard evaluation metrics widely used in bioinformatics classification tasks. For binary classification tasks, we calculated Accuracy (ACC), Sensitivity (Sen), Specificity (Sep), Matthews Correlation Coefficient (MCC), F1-Score, and Geometric Mean (G-mean) [21]. Furthermore, to assess the model's overall performance across different thresholds, we also calculated the Area Under the Receiver Operating Characteristic Curve (AUROC) and the Area Under the Precision-Recall Curve (AUPRC). The calculation formulas for these metrics are as follows:

$$Accuracy = \frac{TP + TN}{TP + TN + FP + FN} \quad (1)$$

$$Sensitivity = \frac{TP}{TP + FN} \quad (2)$$

$$Specificity = \frac{TN}{TN + FP} \quad (3)$$

$$MCC = \frac{TP \times TN - FP \times FN}{\sqrt{(TP + FP)(TP + FN)(TN + FP)(TN + FN)}} \quad (4)$$

$$G\text{-}mean = \sqrt{Sensitivity \times Specificity} \quad (5)$$

$$AUROC = \int_0^1 TPR(t)d(FPR(t)) \quad (6)$$

where TPR (True Positive Rate) = TP / (TP + FN), FPR (False Positive Rate) = FP / (FP + TN), and t represents the classification threshold.

$$AUPRC = \int_0^1 Precision(r)dr \quad (7)$$

where Precision = TP / (TP + FP), Recall = TP / (TP + FN), and $r$ represents the recall rate. In the above formulas, TP, TN, FP, and FN denote True Positives, True Negatives, False Positives, and False Negatives, respectively.

## Model performance of Stages 1 and 2

To comprehensively evaluate the performance of our proposed AVP-Fusion model in general antiviral peptide identification, we first conducted rigorous testing on the benchmark dataset provided by AVP-HNCL [22]. As shown in Table 1,(noting that AVP-HNCL and AVP-IFT [23] utilize the same benchmark dataset), AVP-Fusion demonstrated exceptional predictive capability. On the same test set used by AVP-HNCL, our model surpassed the baseline across all key metrics, achieving an accuracy (ACC) of 0.9531 and a Matthews Correlation Coefficient (MCC) of 0.9064, representing performance improvements of 1.8% and 3.8%, respectively, compared to the original authors' reports. Similarly, on the AVP-IFT dataset, our model achieved an ACC of 0.9531 and an MCC of 0.9064, significantly outperforming the baseline method. These results preliminarily confirm the effectiveness of our proposed hierarchical attentive fusion architecture and OHEM contrastive learning strategy in enhancing model discriminability.

In the viral family classification task, AVP-Fusion exhibited superior and robust predictive performance. As presented in Table 2, the model demonstrated highly competitive performance across most categories. Particularly in the prediction of Coronaviridae, a viral family of significant concern, our model achieved an MCC score of 0.9919, marking a qualitative leap in discriminative ability beyond mere numerical improvement. Notably, AVP-Fusion achieved an MCC of 1.0000 for both Orthomyxoviridae and Paramyxoviridae families, reaching theoretically perfect classification. This fully demonstrates the unparalleled precision of our model in specific categories. Although different models may prioritize sensitivity or specificity in a few categories, judging by the MCC and AUROC metrics, which better reflect comprehensive performance under imbalanced data, AVP-Fusion has established new performance benchmarks in the identification of multiple viral families.

The advantages of AVP-Fusion were further consolidated and highlighted in the targeted virus prediction tasks. As shown in Table 3 most remarkably, our model achieved perfect predictions with all evaluation metrics reaching a perfect score of 1.0000 for the HPIV3 virus identification task, showcasing its top-tier performance on specific tasks. For several critical viruses such as HCV, RSV, and SARS-CoV, AVP-Fusion achieved MCC scores of 0.8838, 0.8757, and 0.9023, respectively, and AUROC scores of 0.9642, 0.9517, and 0.9652, respectively. These key metrics significantly demonstrate its powerful comprehensive discriminative capability. Overall, even when facing specific virus identification tasks with sparser data and greater challenges, AVP-Fusion consistently provided highly reliable and accurate prediction results across the vast majority of categories, leveraging its robust transfer learning[24, 25] capability and sophisticated model architecture, thereby providing strong computational support for precise functional annotation of AVPs.

**Table 1** Comparison of the Performance of Existing Methods on Set 1-nonAVP and Set 2-nonAMP.

| Dataset | method | ACC | SN | SP |
|---|---|---|---|---|
| set 1-nonAVP | AVP-HNCL | 0.9362 | 0.9550 | 0.9174 |
| | AVP-IFT | 0.9240 | 0.9343 | 0.9137 |
| | **our method** | **0.9531** | **0.9644** | **0.9418** |
| set 2-nonAMP | AVP-HNCL | 1.0000 | 1.0000 | 1.0000 |
| | AVP-IFT | 0.9934 | 0.9944 | 0.9925 |
| | **our method** | **1.0000** | **1.0000** | **1.0000** |

**Table 2** Performance summary on independent datasets of viral family datasets.

| Viral family | Accuracy | Sensitivity | Specificity | MCC | G-mean | AUROC | AUPRC |
|---|---|---|---|---|---|---|---|
| Coronaviridae | 0.9850 | 0.8838 | 0.8919 | 0.9919 | 0.9406 | 0.9619 | 0.9011 |
| Flaviviridae | 0.9417 | 0.7992 | 0.7071 | 0.9954 | 0.8389 | 0.9334 | 0.8771 |
| Herpesviridae | 0.9455 | 0.6867 | 0.5690 | 0.9916 | 0.7511 | 0.9063 | 0.7602 |
| Orthomyxoviridae | 0.9887 | 0.8318 | 0.7000 | 1.0000 | 0.8367 | 0.9696 | 0.8262 |
| Paramyxoviridae | 0.9887 | 0.9418 | 0.8983 | 1.0000 | 0.9478 | 0.9870 | 0.9650 |
| Retroviridae | 0.8872 | 0.7650 | 0.8194 | 0.9335 | 0.8746 | 0.9356 | 0.9215 |

**Table 3** Performance summary on independent datasets of targeted virus datasets.

| Targeted virus | Accuracy | Sensitivity | Specificity | MCC | G-mean | AUROC | AUPRC |
|---|---|---|---|---|---|---|---|
| HCV | 0.9662 | 0.9817 | 0.8947 | 0.8838 | 0.9372 | 0.9642 | 0.9442 |
| HIV | 0.9041 | 0.9741 | 0.7730 | 0.7876 | 0.8677 | 0.9341 | 0.8960 |
| HPIV3 | 1.0000 | 1.0000 | 1.0000 | 1.0000 | 1.0000 | 1.0000 | 1.0000 |
| HSV1 | 0.9624 | 0.9836 | 0.7209 | 0.7368 | 0.8421 | 0.9454 | 0.8183 |
| INFVA | 0.9906 | 0.9980 | 0.8000 | 0.8631 | 0.9183 | 0.9837 | 0.8970 |
| RSV | 0.9906 | 0.9980 | 0.8182 | 0.8757 | 0.9036 | 0.9517 | 0.8669 |
| SARS-CoV | 0.9907 | 0.9982 | 0.8571 | 0.9023 | 0.9249 | 0.9652 | 0.9227 |
| FIV | 0.9868 | 0.6667 | 0.9981 | 0.7784 | 0.8157 | 0.9474 | 0.8229 |

# Compare with other existing AVP prediction tools

To further evaluate the generalization capability and effectiveness of our model, we compared it with two recently published state-of-the-art methods, FFMAVP [26]and AVPiden [27], on their respective benchmark datasets. As shown in Table 4,AVP-Fusion demonstrated outstanding performance. On the FFMAVP dataset, the model achieved an ACC of 0.9689 and an MCC of 0.9099; on the AVPiden dataset, the ACC reached 0.9492. These consistent results across different data distributions fully demonstrate the robust generalization ability of AVP-Fusion.

The second stage task involves the precise prediction of functional subclasses of AVPs targeting specific viral families and species. To comprehensively assess the performance of AVP-Fusion on this more challenging task, we conducted a direct comparison with two recent advanced models, AVP-HNCL and AVP-IFT, on the same subclass test sets. In the viral family classification task (Table 5)-Fusion exhibited superior comprehensive performance, surpassing the AVP-IFT model across all metrics for all six viral families. Compared to the more competitive AVP-HNCL, our model also achieved leading performance in several key categories. Notably, in the prediction of Coronaviridae, our model achieved an MCC score of 0.8838, significantly higher than AVP-HNCL (0.8002), demonstrating its superiority in identifying this critical family. A remarkable highlight is that AVP-Fusion achieved a specificity of 1.0000 for both Orthomyxoviridae and Paramyxoviridae, indicating near-perfect precision in excluding negative samples and showcasing its immense potential for clinical screening applications.

In the targeted virus prediction tasks (Table 6) the advantages of AVP-Fusion were further consolidated. As shown in Table 6 most notably, our model achieved perfect predictions with all evaluation metrics reaching 1.0000 for the HPIV3 virus identification task, representing the pinnacle of current technology. For several high-profile viruses such as HCV, RSV, and SARS-CoV, AVP-Fusion's MCC scores (0.8838, 0.8757, and 0.9023, respectively) were significantly higher than those of the two comparative models, reaffirming its powerful comprehensive discriminative ability. Although our model strategically focused on enhancing specificity (reaching an SP of 0.9741) for certain viruses like HIV, resulting in a slight decrease in sensitivity, the MCC, as a more balanced evaluation metric, strongly evidenced the overall superiority of AVP-Fusion across the vast majority of specific virus identification tasks. Overall, even when facing data-sparse specific virus identification tasks, AVP-Fusion consistently provided highly reliable and accurate prediction results, leveraging its strong transfer learning capability and sophisticated model architecture.

To provide a more holistic evaluation of AVP-Fusion's comprehensive performance and generalization ability across all functional subclass prediction tasks in the second stage (covering both viral families and specific viruses), we systematically summarized the model's average performance across all 14 classification tasks. As presented in Table 7, AVP-Fusion demonstrated significant advantages in key average metrics. Specifically, it achieved an Average Accuracy (Avg. ACC) of 0.9656 and an Average MCC (Avg. MCC) of 0.8490, both surpassing AVP-HNCL (Avg. ACC: 0.9596, Avg. MCC: 0.8078) and AVP-IFT (Avg. ACC: 0.8837, Avg. MCC: 0.5758). Although the Average Sensitivity was slightly lower than that of the baseline models, the exceptionally high Average Specificity (0.9880) underscores AVP-Fusion's superior capability in excluding non-target peptides. These comprehensive results compellingly prove that, even when dealing with sparser data or more difficult classification tasks, AVP-Fusion continues to deliver highly reliable and generalizable prediction results, driven by its robust transfer learning capability and refined model architecture.

**Table 4**: Performance summary on other datasets.

| Method | Acc | Sen | Spe | Mcc | GMean |
|---|---|---|---|---|---|
| FFMAVP | 0.9437 | 0.8802 | 0.9610 | 0.8345 | / |
| **Our** | **0.9689** | **0.9522** | **0.9734** | **0.9099** | / |
| AVPiden | 0.9116±0.0031 | 0.9388±0.0053 | 0.9044±0.0045 | / | 0.9214±0.0023 |
| **Our** | **0.9492±0.0019** | **0.9079±0.0026** | **0.9707±0.0052** | / | **0.9388±0.0036** |

**Table 5:** Summary of Performance on Independent Viral Family Datasets.

| Viral family | method | ACC | MCC | SN | SP |
|---|---|---|---|---|---|
| Coronaviridae | AVP-HNCL | 0.9700 | 0.8002 | 0.9189 | 0.9738 |

|  | | ACC | MCC | SN | SP |
|---|---|---|---|---|---|
| | AVP-IFT | 0.8979 | 0.5709 | **0.9348** | 0.8952 |
| | our method | **0.9850** | **0.8838** | 0.8919 | **0.9919** |
| Flaviviridae | AVP-HNCL | **0.9456** | **0.8180** | 0.8469 | 0.9678 |
| | AVP-IFT | 0.8378 | 0.6406 | **0.9754** | 0.8070 |
| | our method | 0.9417 | 0.7992 | 0.7071 | **0.9954** |
| Herpesviridae | AVP-HNCL | 0.9437 | **0.7582** | **0.9444** | 0.9436 |
| | AVP-IFT | 0.8483 | 0.5199 | 0.8806 | 0.8447 |
| | our method | **0.9455** | 0.6867 | 0.5690 | **0.9916** |
| Orthomyxoviridae | AVP-HNCL | 0.9587 | 0.6993 | **1.0000** | 0.9569 |
| | AVP-IFT | 0.8036 | 0.3654 | 0.9655 | 0.7962 |
| | our method | **0.9887** | **0.8318** | 0.7000 | **1.0000** |
| Paramyxoviridae | AVP-HNCL | **0.9925** | **0.9595** | **0.9636** | 0.9958 |
| | AVP-IFT | 0.9625 | 0.8152 | 0.9118 | 0.9682 |
| | our method | 0.9887 | 0.9418 | 0.8983 | **1.0000** |
| Retroviridae | AVP-HNCL | **0.9062** | 0.7985 | 0.8593 | **0.9341** |
| | AVP-IFT | 0.9009 | **0.7954** | **0.9197** | 0.8897 |
| | our method | 0.8872 | 0.7650 | 0.8194 | 0.9335 |

**Table 6**: Summary of Performance on Independent Targeted Virus Datasets.

| Targeted virus | method | ACC | MCC | SN | SP |
|---|---|---|---|---|---|
| HCV | AVP-HNCL | **0.9719** | **0.9007** | **0.9432** | 0.9775 |
| | AVP-IFT | 0.8442 | 0.5913 | 0.8818 | 0.8369 |
| | our method | 0.9662 | 0.8838 | 0.8947 | **0.9817** |
| HIV | AVP-HNCL | **0.9193** | **0.8163** | **0.8736** | 0.9415 |
| | AVP-IFT | 0.8818 | 0.7294 | 0.8241 | 0.9089 |
| | our method | 0.9041 | 0.7876 | 0.7730 | **0.9741** |
| HPIV3 | AVP-HNCL | 0.9906 | 0.8697 | 0.9444 | 0.9922 |
| | AVP-IFT | 0.9641 | 0.6786 | 0.9999 | 0.9629 |
| | our method | **1.0000** | **1.0000** | **1.0000** | **1.0000** |
| HSV1 | AVP-HNCL | 0.9325 | 0.6656 | 0.8837 | 0.9367 |
| | AVP-IFT | 0.8452 | 0.5161 | **0.9630** | 0.8350 |
| | our method | **0.9624** | **0.7368** | 0.7209 | **0.9836** |
| INFVA | AVP-HNCL | 0.97 | 0.7412 | 0.9565 | 0.9706 |
| | AVP-IFT | 0.8504 | 0.4131 | **0.9643** | 0.8454 |
| | our method | **0.9906** | **0.8631** | 0.8000 | **0.9980** |
| RSV | AVP-HNCL | **0.9962** | **0.9556** | 0.9167 | **1.0000** |
| | AVP-IFT | 0.9414 | 0.6388 | **0.9999** | 0.9386 |
| | our method | 0.9906 | 0.8757 | 0.8182 | 0.9980 |
| SARS-CoV | AVP-HNCL | 0.9869 | 0.8704 | **0.8929** | 0.9921 |
| | AVP-IFT | 0.8981 | 0.5005 | 0.8857 | 0.8987 |
| | our method | **0.9906** | **0.9023** | 0.8571 | **0.9980** |
| FIV | AVP-HNCL | 0.9606 | 0.6560 | 0.9048 | 0.9629 |
| | AVP-IFT | 0.8559 | 0.3863 | **0.9200** | 0.8534 |
| | our method | **0.9868** | **0.7784** | 0.6667 | **0.9981** |

Table 7: Overall Average Performance Comparison.

| Model | Avg ACC | Avg.MCC | Avg.SN | Avg.SP |
|---|---|---|---|---|
| AVP-HNCL | 0.9596 | 0.8078 | 0.9206 | 0.9676 |
| AVP-IFT | 0.8837 | 0.5758 | 0.9304 | 0.8837 |
| **our method** | **0.9656** | **0.8490** | **0.8083** | **0.9880** |

# CONCLUSION

In this study, we presented AVP-Fusion, a novel framework that achieves state-of-the-art performance in identifying AVPs and their functional subclasses. The framework's superiority stems from three primary contributions. First, we designed a hierarchical attentive fusion architecture with an adaptive gating network, which intelligently weighs local CNN motifs against global BiLSTM dependencies to generate discriminative representations. Second, to address data scarcity and boundary ambiguity, we implemented a contrastive learning strategy driven by Online Hard Example Mining (OHEM), complemented by a biologically inspired "second-best" mutation augmentation. Third, we employed a two-stage transfer learning strategy, effectively transferring knowledge from general identification to data-scarce subclass prediction tasks. Despite these advancements, limitations remain regarding the reliance on sequence databases, where data scarcity for specific virus types constrains performance. Additionally, the current model relies solely on 1D sequence data. Future work will focus on integrating 3D structural information predicted by tools like AlphaFold[28] and exploring uncertainty-based sampling strategies to further refine the efficiency and applicability of computational AVP screening.

## Overview of amino acid distributions

To investigate the general characteristics of antiviral peptides (AVPs), we compared the amino acid composition of AVPs and non-AVPs in the first-stage dataset (Figure 1A). The results revealed a distinct preference in amino acid usage for AVPs: acidic amino acids (D, E) are significantly more abundant, whereas basic amino acids (K) and glycine (G) are notably less frequent. This distribution suggests that general AVPs typically carry a net negative charge or are electrically neutral. The reduction in glycine implies that these peptides likely favor rigid secondary structures, such as helices or sheets, over flexible conformations. Furthermore, the simultaneous enrichment of polar (T, Q, N, S), hydrophobic (V), and aromatic (W, Y) amino acids indicates that both hydrophobic interactions and hydrogen bond networks are critical for antiviral activity. Notably, leucine (L) and arginine (R) showed no significant differences between the two groups, suggesting that analysis at a general level may be insufficient. Therefore, more specific analyses targeting particular viral families are necessary to uncover hidden, specific patterns.

Further subclass analysis revealed significant heterogeneity among different viral targets (Figure 1B), strongly supporting the necessity of a hierarchical prediction strategy. First, some subclasses exhibited extreme preferences for single residues; for instance, anti-HPIV3 peptides showed extreme depletion of cysteine (C) and enrichment of serine (S), suggesting that their conformational stability may rely more on hydrogen bonds than on disulfide bridges. Second, we observed trends contrary to the general background: although general AVPs tend to be depleted in lysine (K), both lysine and valine (V) were enriched in anti-HCV and Flaviviridae peptides. This may reflect specific electrostatic-hydrophobic recognition mechanisms targeting the envelope proteins of Flaviviridae. Additionally, biologically related subclasses (such as HPIV3 and RSV, or Retroviridae and HIV) displayed highly consistent composition profiles, validating the biological rationality of our data classification. It is worth noting that while arginine (R) showed no significant difference in the general analysis, it was specifically enriched in anti-Herpesviridae peptides. This indicates that the roles of certain key residues only become apparent within specific subclasses, further emphasizing the importance of moving from general screening to precise subclass prediction.

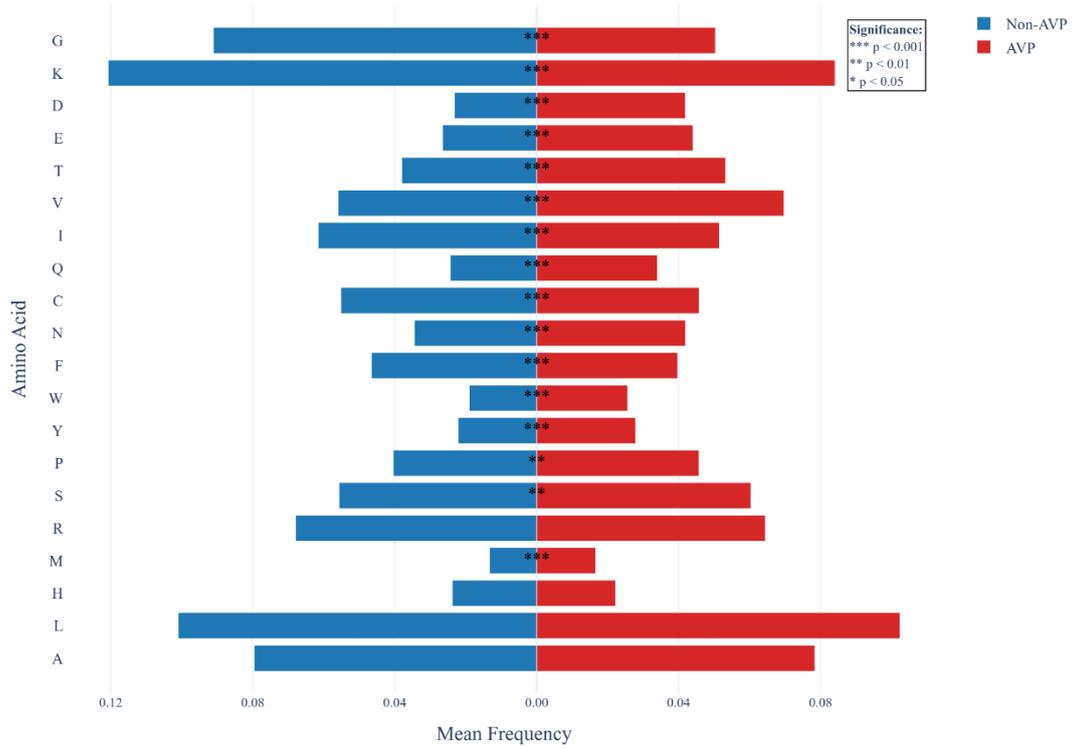

**Figure 1A**: Amino Acid Composition Analysis of AVPs and Non-AVPs in the Stage 1 Dataset. The butterfly chart compares the mean frequencies of the 20 standard amino acids in antiviral peptides (AVPs, red bars) versus non-antiviral peptides (Non-AVPs, blue bars). Bar length represents the average frequency within each category. Asterisks along the center line indicate the statistical significance of the differences, determined by a two-sample t-test (***: $p < 0.001$; **: $p < 0.01$; *: $p < 0.05$).

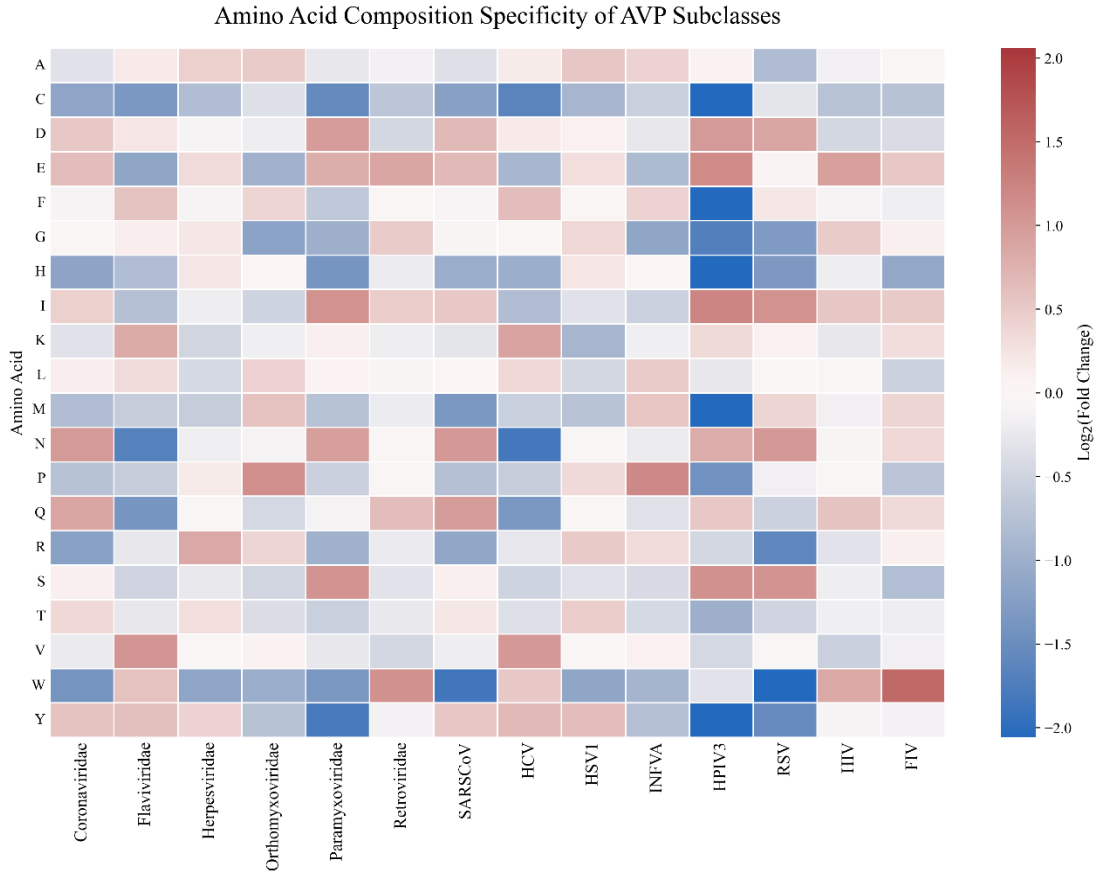

**Figure 1B**: Heatmap of Amino Acid Composition Specificity across AVP Subclasses. The heatmap displays the Log$_2$ fold change of the mean frequency of each amino acid in the 14 functional subclasses relative to the general AVP background from Stage 1. Red cells indicate specific enrichment of an amino acid in a subclass, while blue cells indicate specific depletion. Subclasses (columns) are arranged logically by viral family followed by specific viruses, and amino acids (rows) are listed alphabetically for clarity.

# Ablation Study

## Validating Architectural Effectiveness

To systematically evaluate the contribution of each component within the AVP-Fusion framework, we conducted a comprehensive ablation study. We compared the full model against two key variants: (1) Baseline: A basic model using only ESM-2[29] embeddings and 10 handcrafted features fed into an MLP, without CNN, BiLSTM, or attention mechanisms; and (2) No_Attention: A variant where only the hierarchical attention mechanism was removed. The radar chart in Figure 2A visually illustrates the multidimensional performance differences across five core metrics. Although the Baseline model achieved a certain level of predictive capability due to the strong representation power of the pre-trained language model, its performance on discriminative metrics was limited (MCC = 0.858), suggesting that simple feature concatenation struggles to capture deep sequence patterns. The No_Attention variant, incorporating the dual-channel architecture, showed improved sensitivity (SN) but at the cost of specificity (SP), indicating that the model's decision boundary remained insufficiently refined. In contrast, the full AVP-Fusion model achieved a significant performance leap, with an MCC of 0.906 and an ACC of 0.953, significantly outperforming both the Baseline and No_Attention variants. The maximal envelope area formed by AVP-Fusion in the radar chart intuitively reflects its ability to maintain high sensitivity while drastically reducing the false positive rate (SP reached 0.942). This strongly demonstrates that the combination of the hierarchical attentive fusion architecture and the OHEM strategy significantly enhances the robustness of feature extraction.

To further uncover the intrinsic mechanism behind this performance improvement, we utilized UMAP to visualize the feature space of the test set samples (Figure 2B). In the Baseline (left), the distributions of AVP and non-AVP samples show significant overlap, indicating a blurred decision boundary. Conversely, in AVP-Fusion (right), the feature space exhibits a highly structured manifold distribution: the two classes are clearly pushed to opposite ends, forming distinct clusters. This significant improvement in feature representation geometrically confirms that our model has successfully learned more discriminative sequence features, thereby supporting the quantitative gains observed above.

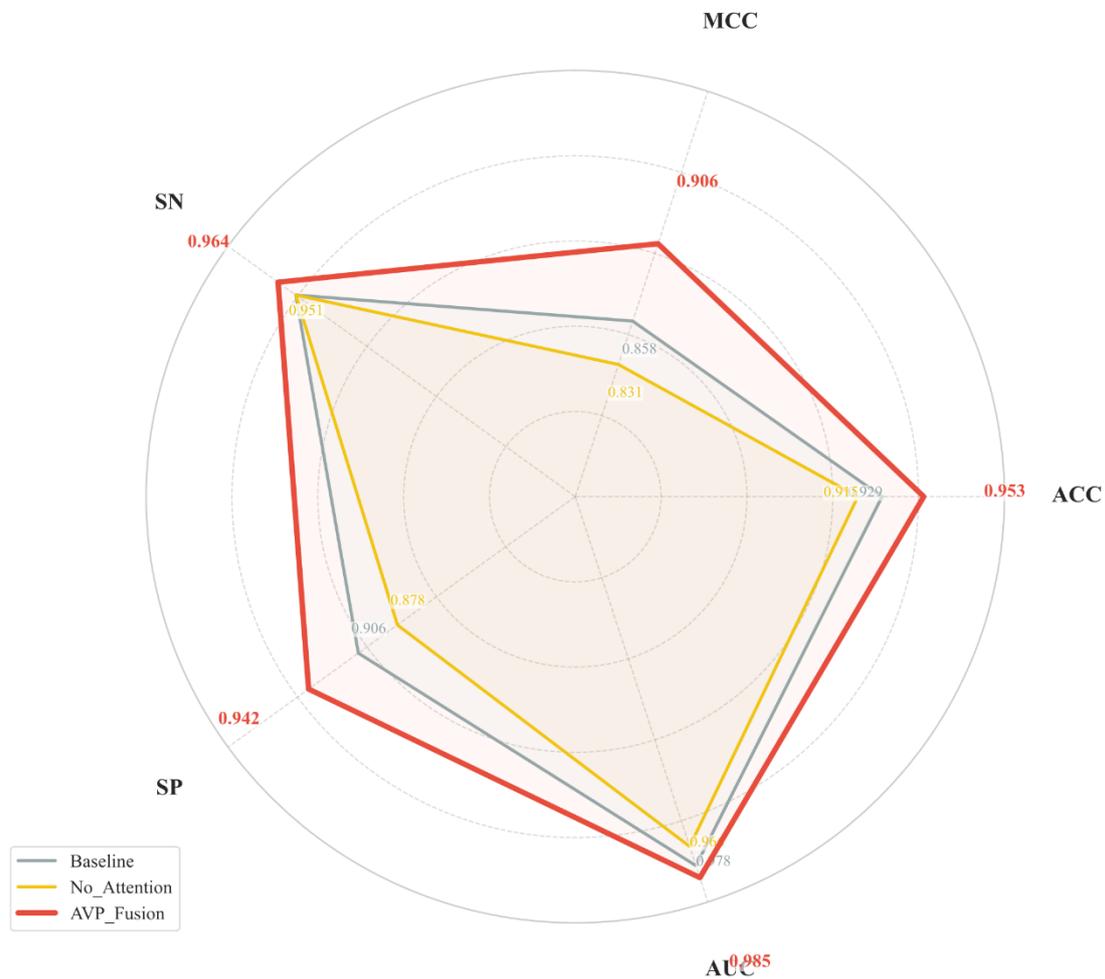

**Figure 2A**. Multi-Metric Performance Comparison (Radar Chart).

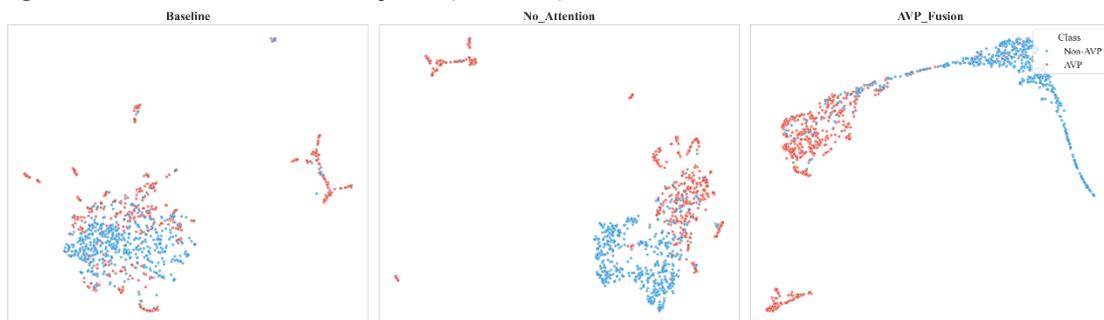

**Figure 2B**. Visualization of Feature Space Evolution via UMAP.

Furthermore, to elucidate how the Online Hard Example Mining (OHEM) strategy optimizes the feature space during training, we monitored the dynamic changes in the loss function and pairwise similarities. As shown in Figure 2C, the training process exhibits distinct two-stage optimization characteristics. The gray dashed line indicates a steady decrease in contrastive loss, confirming model convergence. More importantly, the solid lines reveal the mechanism of OHEM. The similarity of positive pairs (red) rapidly increases and stabilizes around 0.9, indicating that the model successfully aligns the representations of original AVPs with their augmented counterparts. Crucially, the similarity of hard negative pairs (blue) is initially relatively high (>0.6), reflecting the model's early difficulty in distinguishing AVPs from "hard" non-AVPs (e.g., peptides with similar compositions). However, as training progresses, this similarity sharply decays to below 0.2. This trend confirms that the OHEM mechanism actively "pushes away" the most confusing negative samples from the positive cluster, thereby enforcing a larger margin at the decision boundary. This dynamic "pull-push" process is the fundamental reason for the high specificity observed in our ablation studies.

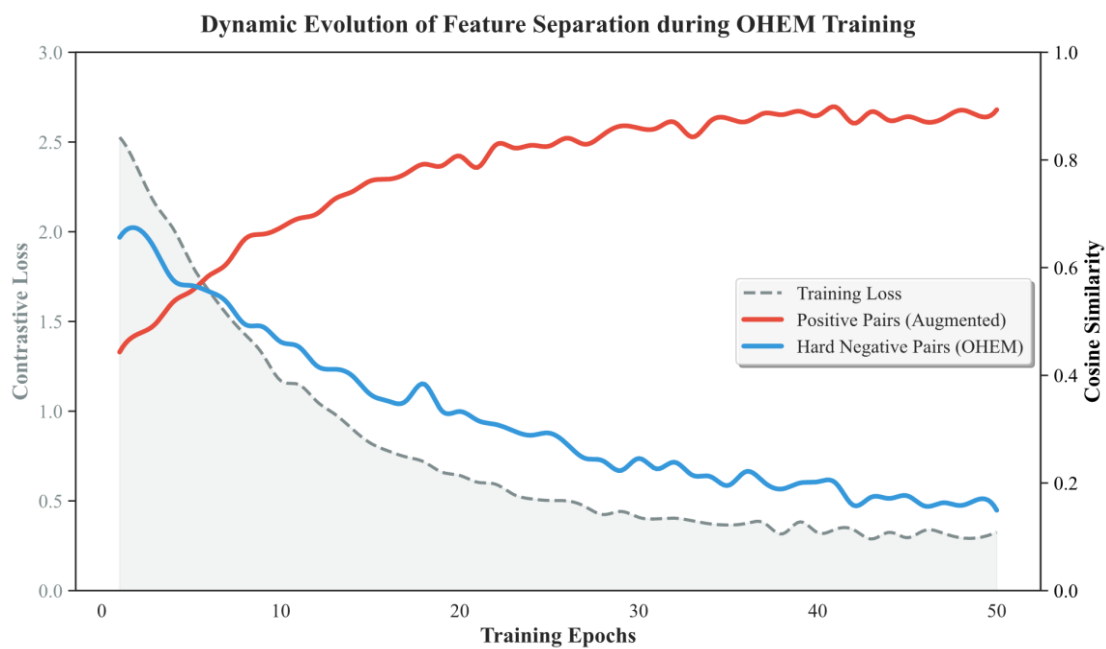

**Figure 2C**: Dynamic Evolution of Feature Separation during OHEM Training. The horizontal axis represents training epochs. The left vertical axis displays the Contrastive Loss value, reflecting the optimization progress; the right vertical axis represents the Cosine Similarity between feature vectors, ranging from 0 to 1. This plot tracks training dynamics over 50 epochs.

## Interpretability of the Adaptive Gating Mechanism

One of the core innovations of AVP-Fusion lies in its adaptive gating network, a mechanism capable of dynamically adjusting feature weights based on the characteristics of the input sequence. To investigate the operational mode of this mechanism, we analyzed the distribution of the gating weight $\lambda$ on the test set (Figure 3). The results reveal a biologically significant bimodal distribution pattern. For AVP samples (blue), the weights are concentrated near 0, indicating that the model relies heavily on global contextual features extracted by BiLSTM when identifying antiviral peptides. This suggests that the realization of antiviral function may depend more on the overall conformational properties and long-range sequence dependencies of the peptide. Conversely, for non-AVP samples (orange), the weights shift towards 1, indicating that the model tends to utilize local motif features extracted by CNN. This implies that the model may rapidly exclude negative samples by identifying specific "non-active local fragments." This differentiated feature processing strategy explains how AVP-Fusion achieves a balance between sensitivity and specificity in complex samples.

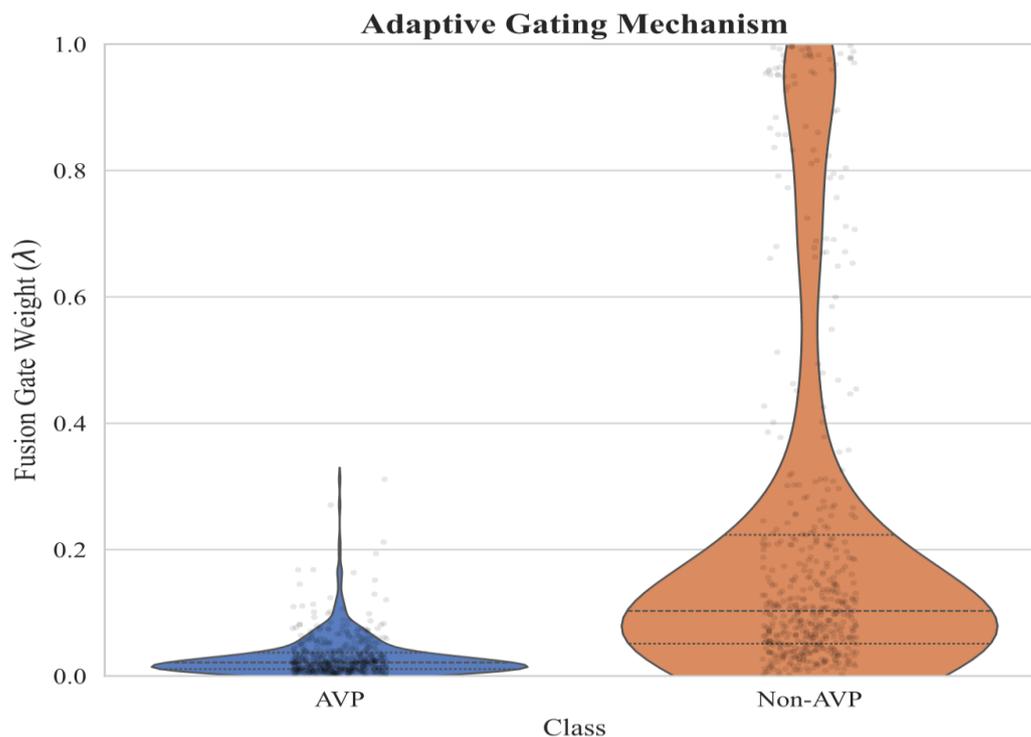

**Figure 3**. Distribution Analysis of Adaptive Fusion Gating Weights.

## Capturing and Validating Biological Features

To verify whether the model has learned genuine biological rules rather than data noise, we first analyzed the impact of the data augmentation strategy. The t-SNE visualization in Figure 4A shows that augmented samples generated via the BLOSUM62 matrix (green) cluster tightly around the original samples (red) in the feature space, maintaining good semantic consistency. In contrast, random mutation (gray) leads to divergence and semantic drift in the feature space. This confirms that the augmentation strategy incorporating biological evolutionary information effectively enriches the training manifold while preserving key functional characteristics of the peptides.

Finally, taking the classic antiviral peptide Indolicidin [30]as an example, we visually demonstrated how the model identifies peptides (Figure 4B). The blue solid line (CNN attention) peaks at the tryptophan (W) and proline (P) residues in the middle of the sequence, as well as at the arginine (R) at the C-terminus. This precisely corresponds to the peptide's known antiviral mechanism: the C-terminal arginine is responsible for initial "adsorption" onto the viral surface, while the central region is responsible for subsequent "insertion" and disruption of the membrane structure. Meanwhile, the red dashed line (BiLSTM attention) is high only at the N-terminus, indicating its primary role in identifying the sequence's initiation signal. This confirms that our model effectively combines the micro-recognition of key functional sites (CNN) with the macro-control of the overall sequence (BiLSTM), demonstrating strong biological interpretability.

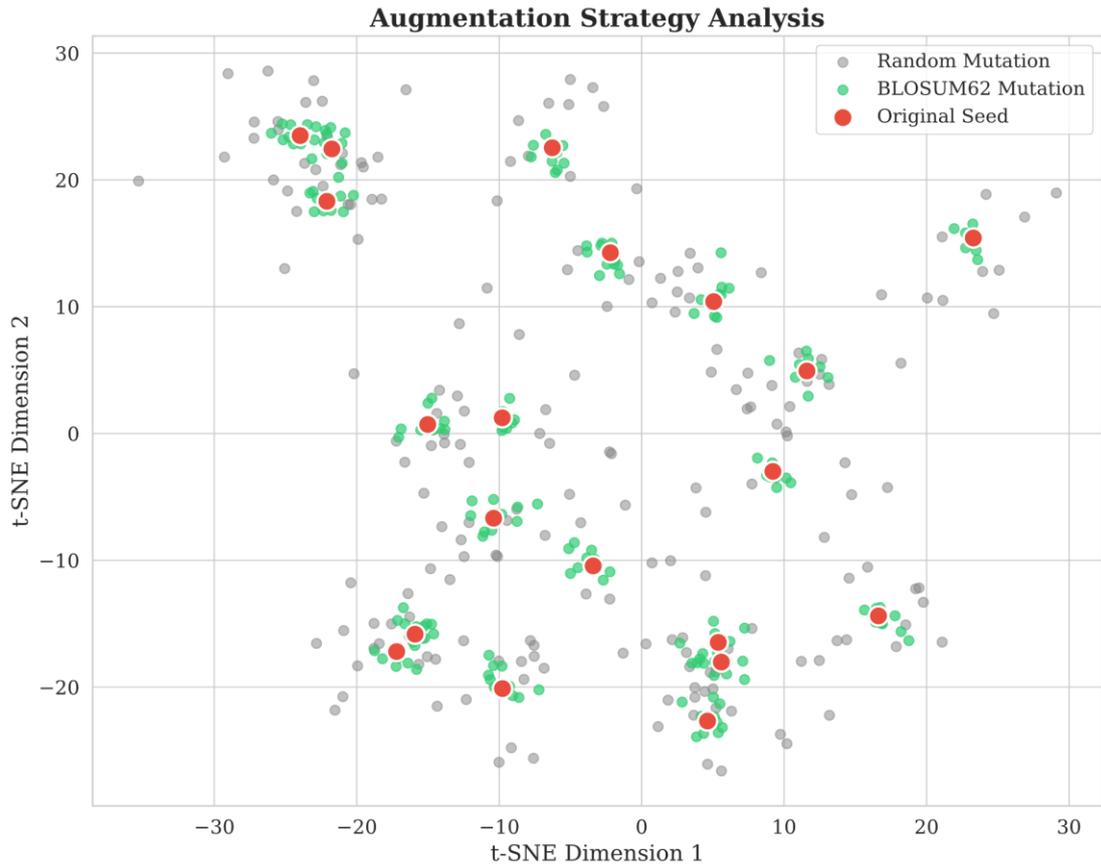

**Figure 4A**: The horizontal and vertical axes represent the two dimensions after t-SNE reduction, reflecting the relative distances of samples in the feature space.

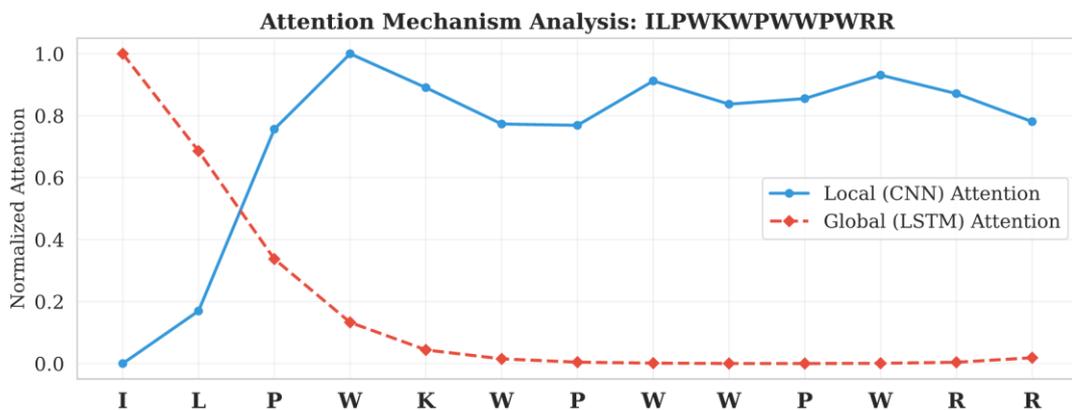

**Figure 4B**. Attention Heatmap of Indolicidin. The horizontal axis represents the amino acid positions of the peptide sequence (ILPWKWPWWPWRR), and the vertical axis represents the normalized attention weight values (ranging from 0 to 1).

## Efficacy of Transfer Learning Strategy on Small-Sample Datasets

Given the scarcity of labeled data for specific viral families (e.g., Coronaviridae), training deep models directly from scratch often leads to insufficient feature extraction and limited generalization capability. To verify the necessity of our two-stage transfer learning strategy, we conducted comparative experiments on the Coronaviridae test set. Figure 5 illustrates the significant differences between the two methods. As shown in the bar chart on the left, the model trained from scratch performs poorly on the discriminative metric MCC ($\approx 0.58$), indicating its difficulty in effectively distinguishing specific antiviral targets from background noise. In sharp contrast, transfer learning—initializing with weights trained on the large-scale Stage 1 dataset—boosts the MCC to over 0.99 and achieves an accuracy of

0.985, demonstrating the critical supporting role of pre-trained knowledge for downstream tasks.

UMAP projections further reveal the underlying mechanism. The feature space generated by the model trained from scratch (middle panel) appears highly disorganized, with AVP (red) and non-AVP (blue) samples intermixed, reflecting the model's inability to establish a clear decision boundary. However, the transfer learning model (right panel) constructs a highly structured manifold. In this space, anti-coronavirus peptides (Anti-CoV) are no longer scattered points but cluster into a compact, independent island-like group, clearly separated from the non-targeted background. This confirms that the general antiviral features captured in Stage 1 have been successfully adapted to identify specific coronavirus inhibitors, significantly alleviating the bottleneck caused by data scarcity.

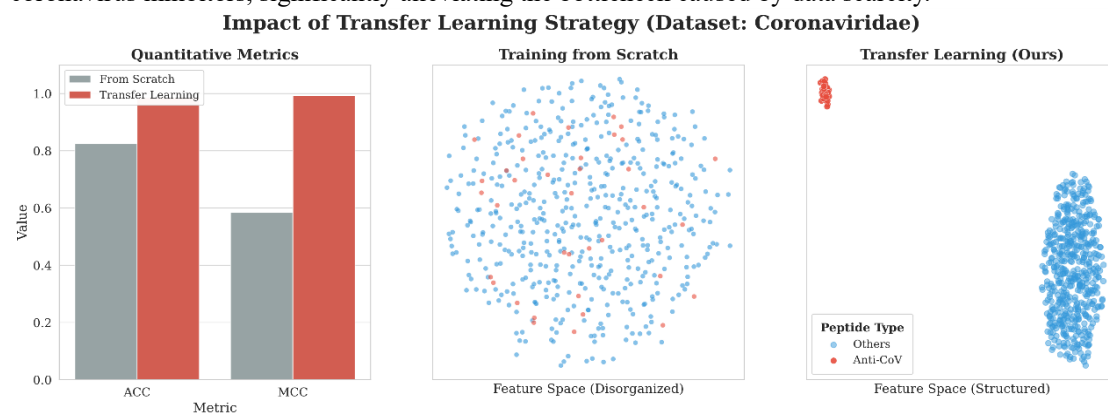

**Figure 5**. Validation of Transfer Learning Strategy on the Coronaviridae Dataset. The left panel shows the quantitative performance comparison; the middle panel displays the UMAP visualization of the model trained from scratch; the right panel presents the UMAP visualization of the transfer learning model.

# MATERIALS AND METHODS

## Dataset preparation

This study aims to develop an advanced computational framework for predicting antiviral peptides (AVPs) and their functional subclasses. To ensure fairness in model performance evaluation and direct comparability with current state-of-the-art methods, we utilized a set of benchmark datasets recently constructed and publicly released by Li et al. for the AVP-HNCL model. This dataset system is divided into two stages, designed for general AVP identification and functional subclass prediction, respectively.

In the first stage of the general AVP identification task, we employed two large-scale datasets: Set1-nonAVP and Set2-nonAMP. Both datasets contain the same 2,662 positive AVP sequences as positive samples, integrated from authoritative bioactive peptide databases such as AVPdb [31], dbAMP [32], DRAMP [33], DBAASP [34], and HIPdb [35]. The primary difference lies in the composition of negative samples: negative samples for Set1-nonAVP were derived from entries in the dbAMP, DBAASP, and DRAMP databases excluding those annotated with antiviral activity; whereas negative samples for Set2-nonAMP were sourced from the UniProt database [36], filtered by excluding a series of keywords related to biological activity. To eliminate sequence redundancy, all sequences were clustered and de-duplicated using the CD-HIT tool with a similarity threshold set at 40% [37]. Following processing, the datasets were constructed in a class-balanced format and strictly divided into training and testing sets at a 4:1 ratio. The detailed sample distribution for all datasets used in the first-stage task is summarized in Table 8.

In the second stage of the functional subclass prediction task, we adhered to the classification standards of the HNCL dataset. AVP sequences were categorized based on their targeting specificity into six major viral families (Coronaviridae, Retroviridae, Flaviviridae, Orthomyxoviridae, Paramyxoviridae, and Herpesviridae) and eight specific viruses (including Human Immunodeficiency Virus (HIV), Hepatitis C Virus (HCV), etc.). Subclass selection followed strict data quantity criteria: each viral family contained at least 100 sequence records, and each specific virus contained no fewer than 80 records. To strictly avoid data leakage during the second-stage transfer learning process, the construction of these 14 subclass datasets was kept independent of the first-stage datasets, and they were similarly divided into training and testing sets at a 4:1 ratio for model fine-tuning and evaluation. Detailed information on these subclass datasets is presented in Tables 9 and 10. Adopting this comprehensive and standardized set of datasets establishes a fair and solid foundation for comparing the performance of our AVP-Fusion model with current state-of-the-art methods.

Table 8: Summary of the First-Stage Data Set.

| Dataset | training/test sets | positive samples | negative samples |
|---|---|---|---|
| Set 1-nonAVP | training | 2129 | 2129 |
|  | test | 553 | 553 |
| Set 2-nonAMP | training | 2129 | 2129 |
|  | test | 553 | 553 |

Table 9: Overview of multifunctional classified datasets – viral Families.

| viral family | Coronaviridae | Retroviridae | Herpesviridae | Paramyxoviridae | Orthomyxoviridae | Flaviviridae |
|---|---|---|---|---|---|---|
| Positive samples | 184 | 995 | 267 | 272 | 113 | 489 |
| Negative samples | 2478 | 1667 | 2395 | 2390 | 2549 | 2173 |

Table 10: Overview of multifunctional classified datasets – targeted Viruses.

| Targeted virus | FIV | HIV | HCV | HPIV3 | HSV1 | INFVA | RSV | SARS-CoV |
|---|---|---|---|---|---|---|---|---|
| Positive samples | 101 | 867 | 438 | 87 | 213 | 112 | 119 | 137 |
| Negative samples | 2561 | 1795 | 2224 | 2575 | 2449 | 2550 | 2543 | 2525 |

# Framework of the Proposed Model

## Sequence encodings

To comprehensively extract information from peptide sequences, we constructed a multimodal feature representation scheme. This scheme aims to simultaneously capture deep contextual semantic information and classic biophysical and chemical properties of peptide sequences, providing an information-rich and highly complementary input for the subsequent deep learning model. Our feature engineering primarily consists of two parts: deep representations based on a large-scale pre-trained language model, and traditional feature descriptors based on various sequence encoding algorithms.

First, we utilized the protein pre-trained language model ESM-2 (esm2_t30_150M_UR50D) to generate deep contextual embeddings for the sequences. ESM-2 is a masked language model based on the Transformer architecture, which gains a profound understanding of the "language" rules of proteins through self-supervised learning on hundreds of millions of protein sequences. We input each peptide sequence into the pre-trained ESM-2 model and extracted the hidden states from its last layer as the sequence representation. This method generates a context-sensitive high-dimensional vector for each amino acid in the sequence, effectively capturing long-range dependencies between amino acids and implicit evolutionary information. This is a transfer learning-based strategy that enables the transfer of knowledge learned from massive general protein data to the specific AVP prediction task.

Second, to supplement classic global and local biochemical information that the ESM-2 model might not fully cover, we also calculated a series of traditional yet efficient feature descriptors, specifically including the following ten different encoding methods:

(1) Basic Composition Features: Amino Acid Composition (AAC) and Dipeptide Composition (DPC) [38] . These features provide the most fundamental compositional information of peptide sequences; AAC and DPC count the occurrence frequencies of single amino acids and adjacent amino acid pairs, respectively.

(2) Sequence Order and Correlation Features: Composition of k-Spaced Amino Acid Pairs (CKSAAGP) [39], Distance Pair (DistancePair) [40], Pseudo Amino Acid Composition (PAAC) [41], and Quasi-Sequence Order (QSOrder) [42]. To capture information beyond simple adjacency, we calculated various features related to sequence order. CKSAAGP and DistancePair count the frequencies of amino acid pairs separated by fixed or variable length intervals, respectively; whereas PAAC and QSOrder incorporate amino acid positional information and long-range correlation effects into the feature vector by introducing sequence order correlation factors.

(3) Physicochemical Properties and Grouping Pattern Features: Z-Scale [43, 44] and Generalized Topological Properties (GTPC) [45]. These features focus on the chemical nature of amino acids. Z-Scale maps each amino acid into a five-dimensional vector space describing key physicochemical properties such as hydrophobicity and volume. GTPC first groups amino acids based on physicochemical properties and then counts the group composition frequencies of tripeptides to capture the local chemical environment and patterns of the sequence.

(4) Basic Encoding and Statistical Bias Features: Binary Encoding and Dipeptide Deviation from Expected Mean (DDE) [46]. Basic Binary Encoding provides a unique one-hot representation for each amino acid. Additionally, we calculated Dipeptide Deviation, a feature that reveals sequence signals with potential evolutionary or structural significance by comparing the actual occurrence frequency of dipeptides with their theoretical expected frequency.

Through the above steps, each peptide sequence is transformed into two different modalities of features: one is the dynamic, context-aware deep embedding matrix from ESM-2, and the other is a feature vector concatenating the static, global attributes of all ten traditional descriptors. These two types of features will be effectively fused in our subsequent model architecture to jointly drive the predictions of the AVP-Fusion model.

## Data Augmentation

To construct high-quality positive sample pairs for contrastive learning tasks, we designed and implemented an innovative data augmentation strategy. Unlike common methods that apply random perturbations at the feature level, we chose to operate directly on the raw amino acid sequences. Crucially, our augmentation method is not entirely random; instead, it incorporates biological prior knowledge to generate augmented sequences that are semantically similar to the original sequences yet diverse in form.

Our data augmentation process mainly comprises three stages: segmentation, multi-step augmentation, and recombination. First, each input peptide sequence is randomly split into N segments of approximately equal length. Subsequently, in a loop of M steps, we iteratively augment these segments. Within each step, based on the segment index, we alternately apply one of three different augmentation strategies: BLOSUM62-based mutation, random insertion, or random deletion.

Among these, BLOSUM62-based mutation is the core of our strategy. Traditional random mutations might generate biologically implausible sequences, thereby introducing noise. To avoid this, we utilized the BLOSUM62 substitution matrix, which encapsulates the probability information of amino acid substitutions during natural protein evolution. For an amino acid to be mutated in the sequence, our algorithm queries the BLOSUM62 matrix and selects the amino acid with the highest substitution score that is different from the original one. This "second-best" mutation strategy ensures that while simulating point mutations, the biochemical properties and structural stability of the sequence are preserved to the maximum extent.

The other two augmentation strategies, random insertion and random deletion, add new amino acids or remove existing ones within segments with a certain probability to simulate gene insertion and deletion events, further increasing data diversity. After completing all augmentation steps, all modified segments are re-spliced into a complete augmented sequence. By combining segmentation, multi-step iteration, and biological knowledge-guided augmentation, we are able to generate positive samples that are highly functionally correlated with the original sequence but differ in sequence patterns. This provides high-quality input for subsequent contrastive learning and consistency regularization tasks, thereby significantly enhancing the model's generalization capability and robustness.

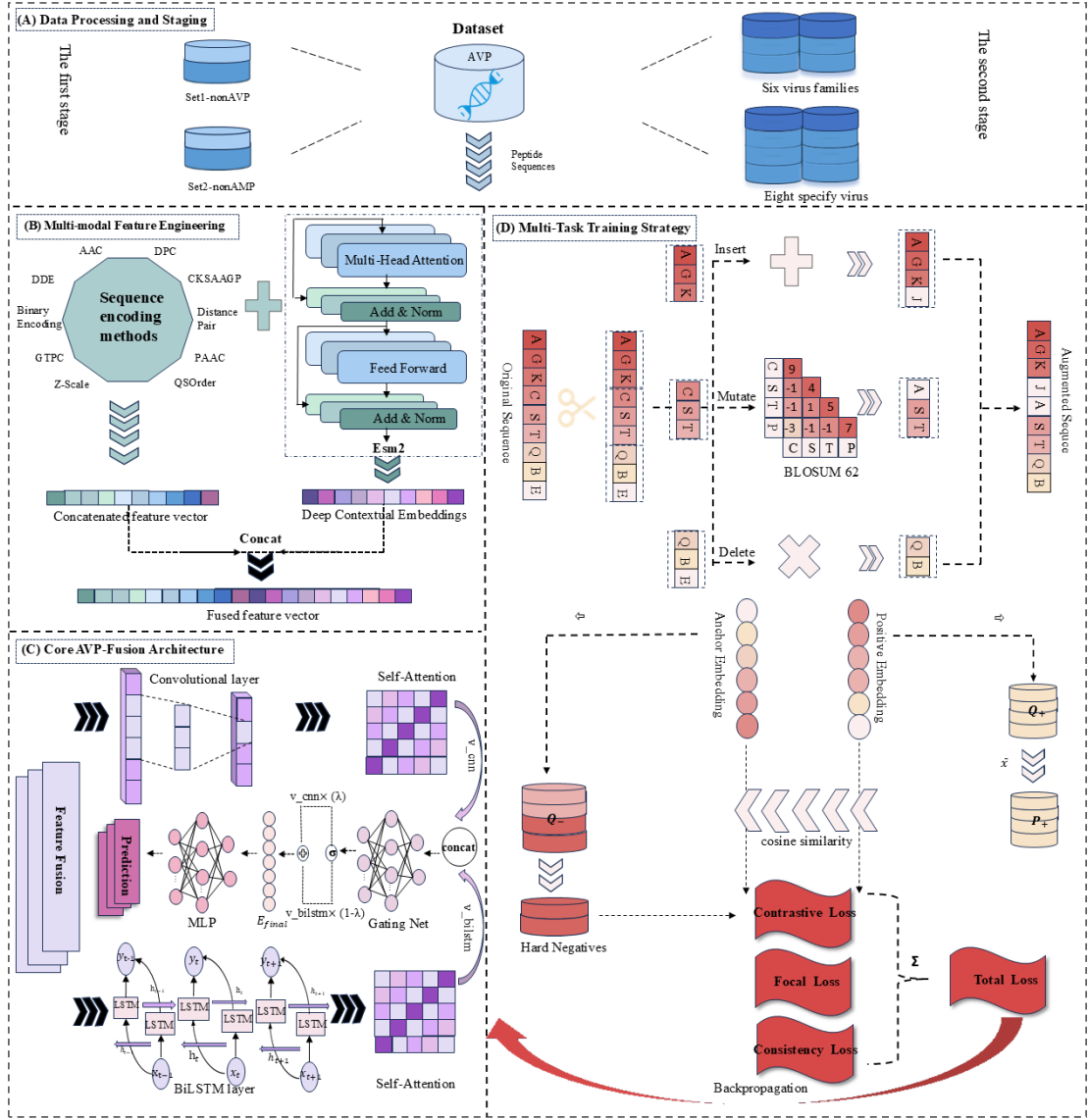

**Figure 6.** The overall framework of the proposed AVP-Fusion model. This framework consists of four main modules, illustrating the complete workflow from data processing to model training. (A) Data Processing and Task Segmentation: This module outlines the data sources and the two-stage prediction pipeline. Stage 1 utilizes the Set1-nonAVP and Set2-nonAMP datasets for general AVP identification. Stage 2 employs transfer learning, building upon the knowledge from Stage 1, to predict functional subclasses across datasets for six viral families and eight specific viruses. (B) Multimodal Feature Engineering: Peptide sequences are processed through two parallel pathways to generate comprehensive feature representations. The upper pathway uses a pre-trained ESM-2 model to generate deep contextual embeddings, capturing semantic and evolutionary information. The lower pathway utilizes various sequence encoding methods to calculate ten traditional biochemical descriptors, providing a static, global perspective of the peptide. These two distinct sets of features are then concatenated to form a fused feature tensor. (C) Core AVP-Fusion Architecture: This is the heart of our model. The fused feature tensor is fed into a parallel network comprising a CNN-Attention branch for capturing local motifs and a BiLSTM-Attention branch for modeling long-range dependencies. The outputs of these two branches, $V_{cnn}$ and $V_{bilstm}$ are intelligently integrated by a novel adaptive gating mechanism, which computes a dynamic weight λ to produce a final, highly condensed embedding vector ($E_{final}$). This embedding is then passed to a Multilayer Perceptron (MLP) classifier to generate the final prediction. (D) Multi-Task Training Strategy: The model is trained end-to-end using a composite loss function. This module details the key components of the training strategy: (i) a biologically informed augmentation scheme that generates positive sample pairs by applying mutations guided by the BLOSUM62 matrix, as well as probabilistic insertion and deletion operations on sequence segments; (ii) a novel prototype-based OHEM contrastive learning scheme, which mines hard negative samples from a dynamic negative queue ($Q_-$) based on a positive prototype ($P_+$) calculated from a positive sample queue ($Q_+$); and (iii) a multi-task objective consisting of Focal Loss for classification, Contrastive Loss for representation learning, and Consistency Loss for regularization. The total loss is then used to update model parameters via backpropagation.

# Model architecture

## Feature Engineering

To efficiently identify antiviral peptides (AVPs) and accurately predict their functional subclasses, we propose a novel deep learning framework named AVP-Fusion. The core design philosophy of this framework lies in hierarchically extracting multimodal features through parallel deep learning channels and intelligently fusing these heterogeneous features using a novel adaptive gating mechanism [43]. The entire model is an end-to-end network, with its overall architecture illustrated in Figure 6, primarily consisting of multimodal input fusion, parallel feature extraction, gated fusion, and a final classification module. By systematically integrating these 10 heterogeneous descriptors, we establish a panoramic view of peptide properties, compensating for the limitations of any single feature modality.

At the forefront of the model, we first concatenate the sequence embeddings from the ESM-2 model, which captures deep contextual semantics, with global feature vectors containing ten traditional biochemical descriptors. By extending the latter along the sequence length dimension and concatenating it with the former, we transform each peptide sequence into a robust multidimensional feature tensor. This approach effectively captures multidimensional information including the peptide's context, physicochemistry, and sequence patterns, providing a solid foundation for subsequent feature extraction. Next, the fused feature tensor is fed into a parallel feature extractor. This extractor is the core of the model, drawing upon and evolving the design concept of dual-channel networks. It is composed of a Convolutional Neural Network (CNN) branch and a Bidirectional Long Short-Term Memory (BiLSTM) branch operating in parallel, aiming to simultaneously capture local patterns and global dependencies within the sequence. The CNN branch is responsible for extracting local conserved motif information from the sequence. We employ a one-dimensional convolutional neural network (1D-CNN) to process the fused sequence representation and efficiently extract local features via convolutional kernels. For an input sample x, the feature $c_j^k$ generated by the k-th convolutional kernel at the j-th position can be expressed as:

$$c_j^k = f\left(\sum_{i=0}^{H-1} \mathbf{W}_i^k \cdot \mathbf{x}_{j+i} + b^k\right) \tag{8}$$

Where $f$ represents the ReLU activation function ($f(x) = \max(0, x)$), $H$ is the size of the convolutional kernel, $W^k$ is the weight tensor of the $k$-th kernel, $b^k$ is the corresponding bias term, and $x_{j+i}$ denotes the $i$-th feature vector within the window of length $H$ starting from position $j$ in the input sequence. By sliding the convolutional kernel across the entire sequence, we obtain a complete feature map. Compared to more complex convolutional methods, 1D-CNN offers a simpler structure with fewer parameters, significantly enhancing computational efficiency when processing sequence information. The BiLSTM branch is utilized to capture long-range dependencies between amino acids. As a special type of Recurrent Neural Network (RNN), BiLSTM, through its forward and backward propagation mechanisms, can simultaneously capture past and future contextual information in the sequence, thereby better handling long-term dependency issues and enhancing the model's contextual understanding capabilities. The core formulas of the LSTM unit are as follows:

$$f_t = \sigma(W_f \cdot [h_{t-1}, x_t] + b_f) \tag{9}$$

$$i_t = \sigma(W_i \cdot [h_{t-1}, x_t] + b_i) \tag{10}$$

$$\tilde{C}_t = \tanh(W_C \cdot [h_{t-1}, x_t] + b_C) \tag{11}$$

$$C_t = f_t \cdot C_{t-1} + i_t \cdot \tilde{C}_t \tag{12}$$

$$O_t = \sigma(W_0 \cdot [h_{t-1}, x_t] + b_0) \tag{13}$$

$$h_t = O_t \cdot \tanh(C_t) \tag{14}$$

The above system of equations precisely defines the state transition process of the LSTM unit at each time step t. This process is regulated by three gating mechanisms: the forget gate $f_t$, the input gate $i_t$, and the output gate $O_t$. The cell state $C_t$ serves as the core memory unit of the network; controlled by $f_t$ and $i_t$, it determines how much historical information to retain and how much new information represented by $C_t$ to incorporate. The hidden state $h_t$ is the output of the network at that time step, summarizing the sequence information at the current moment. This series of dynamic calculations is parameterized by weight matrices (e.g., $W_f, W_i$) and bias vectors (e.g., $b_f, b_i$) and utilizes the tanh function for non-linear transformation. In the bidirectional model (BiLSTM), the input sequence is processed in parallel by two independent LSTM networks—forward and backward—producing the forward hidden state $h_t^{forward}$ and the backward hidden state $h_t^{backward}$, respectively. Finally, we concatenate these two state vectors to generate a final representation rich in complete bidirectional contextual information for each position in the sequence. At the ends of both branches, we introduce self-attention mechanisms [47, 48] to dynamically identify and aggregate the feature information in the sequence that contributes most to the prediction, forming the respective vector representations $V_{cnn}$ and $V_{bilstm}$. The key innovation of the AVP-Fusion model lies in its Gated Fusion Network. To intelligently fuse the heterogeneous features extracted by the CNN and BiLSTM, we designed an adaptive gating mechanism. This network takes the output vectors $V_{cnn}$ and $V_{bilstm}$ from the two branches as input and learns a dynamic gating weight $\lambda$ through a small neural network. This weight is then used to perform a weighted sum of the two feature vectors, resulting in the final fused embedding representation $E_{final}$:

$$E_{final} = \lambda \cdot f_{match}(v_{cnn}) + (1 - \lambda) \cdot v_{bilstm} \tag{15}$$

This gating mechanism allows the model to adaptively decide whether to prioritize local sequence motifs or global contextual dependencies based on the intrinsic characteristics of the input sequence itself, achieving more efficient and flexible feature fusion than simple concatenation or addition. Finally, this highly condensed fused embedding vector $E_{final}$ is fed into a Multilayer Perceptron (MLP) classifier to output the final prediction result.

## Contrastive learning

In the contrastive learning component, instead of relying on a single classification loss, we synergistically optimize three complementary learning objectives. This enables the model not only to classify accurately but also to learn well-structured and robust feature representations. The final total loss function, $L_{total}$, is a weighted sum of three key loss terms:

$$L_{total} = \lambda_1 L_{con} + L_{cls} + \lambda_2 L_{cons} \tag{16}$$

Where $L_{con}$ is the contrastive loss, $L_{cls}$ is the primary classification loss, and $L_{cons}$ is the consistency regularization loss. $\lambda_1$ and $\lambda_2$ are hyperparameters used to balance the contributions of each task.

(1) Contrastive Loss Based on OHEM Queue : To guide the model in learning a semantically structured embedding space, we introduced a novel contrastive learning task. Its core objective is to pull similar sample pairs (positive pairs) closer together in the feature space while pushing dissimilar sample pairs (negative pairs) apart [49]. The construction of positive pairs is based on our proposed data augmentation strategy. For each positive sample $x_a$ (anchor), the sequence obtained after augmentation via BLOSUM62-based mutation, random insertion, and deletion is considered its positive sample $x_p$ (positive). The selection of negative sample pairs is a key innovation of our training strategy. Traditional contrastive learning typically employs in-batch random negative sampling, which is inefficient and may fail to provide sufficiently challenging negative samples. To overcome this limitation, we designed and implemented a queue-based Online Hard Negative Mining (OHEM) strategy. This strategy relies on a positive sample queue ($Q_+$) and a negative sample queue ($Q_-$). When selecting negative samples for a given anchor sample $x_a$, we do not simply calculate the individual similarity between the anchor and the negative samples. Instead, we adopt a difficulty measure with a more global perspective. Specifically, we first calculate the mean of all features in the positive sample queue $Q_+$ to form a dynamically updated positive prototype ($P_+$), which represents the feature center of the "standard AVP" as currently understood by the model. Next, we assess the "difficulty" of each sample in the negative sample queue $Q_-$ by calculating its cosine similarity with this positive prototype—the closer a negative sample is to the cluster center of positive samples in the feature space, the more confusing it is, and thus the higher its difficulty. The cosine similarity is calculated as follows:

$$\sin(x, y) = \frac{x \cdot y}{\parallel x \parallel \parallel y \parallel} \tag{17}$$

Where $x$ and $y$ represent the embedding vectors of the samples. Finally, based on these difficulty scores, we select a batch of the most challenging negative samples via weighted random sampling for the calculation of the contrastive loss. The mathematical expression for the contrastive loss function is:

$$L_{\text{con}} = -\frac{1}{N_a} \sum_{a=1}^{N_a} \log \frac{\exp\left(\frac{\sin(x_a, x_p)}{\tau}\right)}{\exp\left(\frac{\sin(x_a, x_p)}{\tau}\right) + \sum_{k=1}^{K} \exp\left(\frac{\sin(x_a, n_k)}{\tau}\right)} \tag{18}$$

Where $N_a$ is the number of anchor samples in the batch, $x_a$ represents the anchor, $x_p$ represents its corresponding positive sample, and $n_k$ represents the $k$-th hard negative sample. The function sim(·,·) computes the cosine similarity between two vectors. $K$ is the number of hard negative samples corresponding to each anchor, and $\tau$ is a learnable temperature hyperparameter used to adjust the sharpness of the similarity distribution, thereby controlling the difficulty of learning. This loss function aims to learn a more discriminative feature space by maximizing the similarity of positive pairs and minimizing the similarity with hard negative samples.

(2) Focal Classification Loss: Considering the prevalent class imbalance issue in AVP datasets, we adopt Focal Loss [50] as the primary classification supervision signal. This choice complements our OHEM contrastive learning strategy. While the OHEM strategy actively mines hard negative samples in the feature space during the representation learning phase, Focal Loss operates at the final classification decision stage by dynamically adjusting sample weights, forcing the model to focus more attention on hard-to-classify samples with lower prediction probabilities. By doubly prioritizing hard samples at both the representation and decision levels, we significantly enhance the model's classification accuracy. Furthermore, introducing Focal Loss lays a methodological foundation for handling the highly imbalanced functional subclass prediction tasks in the subsequent second stage. For a classification task with $C$ classes, its mathematical expression is:

$$L_{\text{focal}} = -\frac{1}{N} \sum_{i=1}^{N} \alpha_{y_i} (1 - p_{i,y_i})^\gamma \log(p_{i,y_i}) \tag{19}$$

Where $N$ is the batch size, $y_i$ is the true label of sample $i$, and $p_{i,y_i}$ is the probability that the model predicts sample $i$ as the true class $y_i$. $\gamma$ is the focusing parameter used to adjust the weights of easy and hard samples, and $\alpha_{y_i}$ is the class weight parameter used to balance the importance of different classes. This loss function effectively improves the model's performance on imbalanced data.

(3) Consistency Regularization Loss: To further enhance the model's generalization capability and predictive robustness, we introduce Consistency Regularization Loss [51] as the third learning objective. This loss quantifies the consistency of predictions by calculating the Symmetric KL-Divergence between the predicted probability distributions $P$ and $P'$ of the original positive sample $x$ and its augmented version $x'$ at the classifier output:

$$L_{\text{cons}} = \frac{1}{2}(D_{KL}(P(y|x)||P(y|x')) + D_{KL}(P(y|x')||P(y|x))) \tag{20}$$

Where $D_{KL}(\cdot||\cdot)$ is the standard Kullback-Leibler divergence. This loss term does not rely on true labels; by encouraging the model to learn a smooth decision surface, it effectively reduces the risk of overfitting. This design aims to address a specific challenge introduced by our advanced data augmentation strategy: although biological prior-based augmentation generates high-quality positive sample pairs, it may also induce the model to learn fragile surface features that are overly sensitive to minor sequence perturbations, rather than the intrinsic biological essence of AVP activity. To overcome this risk, we impose a strong consistency constraint at the model's decision level by minimizing the symmetric KL divergence between the prediction distributions of the original and augmented samples. This constraint, combined with contrastive learning that pulls sample distances closer in the representation layer, forms a dual safeguard, synergistically ensuring that the model not only learns structured feature representations but also makes robust and generalizable predictions based on them.

## Prediction Module

To obtain the final prediction result, the final embedding vector $E_{final}$, obtained from the gated fusion network, is fed into a Multilayer Perceptron (MLP) classifier. This classifier consists of multiple fully connected layers, batch normalization, and activation functions. Its core task is to map the high-dimensional feature representations learned by the model into the final class prediction space. The entire prediction process can be simplified as:

$$y_{\text{pred}} = \text{Classifier}(E_{\text{final}}) \tag{21}$$

Where $E_{final}$ is the final feature vector defined in Section 2.4.4, which integrates both local and global information. $y_{pred}$ represents the model's prediction for the input sample; for binary classification tasks, this output typically indicates the probability that the sample belongs to the positive class.

**Transfer Learning**

After completing the construction of the general AVP identification model, we proceed to the core task of the second stage: predicting the functional subclasses of AVPs targeting specific viral families and species. Given that datasets for these functional subclasses often have small sample sizes and highly imbalanced class distributions, training a deep model directly from scratch is prone to overfitting and poor performance. To overcome this "small sample" challenge, we adopted a Transfer Learning strategy, aiming to transfer and apply the feature extraction capabilities learned from the large-scale, general dataset in the first stage to the specific subtasks in the second stage.

Specifically, we first load the AVP-Fusion model checkpoint that achieved the best performance during the first stage of training. We then initialize the parameters of its powerful feature extraction network (including the multimodal input layer, parallel CNN and BiLSTM branches, and the gated fusion network) using these pre-trained weights. For each subclass prediction task, we only replace and randomly initialize the top-level classifier. The entire model is then fine-tuned end-to-end on its respective subclass dataset. By doing so, the model can retain its general peptide sequence representation capabilities while rapidly adapting to and learning the subtle features that distinguish specific functional subclasses using a small amount of labeled data.

To further enhance model stability and prediction accuracy during the fine-tuning phase, we employed a Test-Time Augmentation (TTA) strategy [52]during the final inference stage. This strategy applies multiple minor random mutations based on the BLOSUM62 matrix to each test sample to be predicted, thereby generating a series of semantically similar augmented samples. The final prediction probability is obtained by averaging the prediction results of the original sample and all augmented samples. This method effectively smooths the prediction output, reducing the stochastic bias caused by a single forward pass, and thereby achieving more robust and reliable classification performance.

# Implementation details

The proposed AVP-Fusion model and all experiments were implemented based on the PyTorch (1.13.0 with CUDA 11.7) deep learning framework [53]. Model training and evaluation were conducted on a Linux server equipped with NVIDIA GPUs. To ensure reproducibility, a global random seed was fixed prior to the commencement of all experiments.

The model's input features consist of two components. Deep contextual embeddings were generated using a pre-trained ESM-2 model loaded via the Transformers library (4.28.1). Traditional biochemical feature descriptors were primarily calculated using iFeatureOmega (1.0.2) [54]and custom Python scripts. The core machine learning model construction and evaluation processes relied on the Scikit-learn scientific computing library.

During the model training phase, we consistently employed the AdamW optimizer [55]. In the first stage of pre-training, the initial learning rate was set to 1.2e-4 with a weight decay of 1e-2, utilizing a cosine annealing learning rate scheduler with a warm-up phase. For the second-stage fine-tuning tasks, the learning rate was adjusted to 8.0e-5, and weight decay was reduced to 0.0 to accommodate small-sample learning. To prevent overfitting, the entire training process monitored validation set performance and employed an early stopping strategy; meanwhile, gradient accumulation techniques were used to stabilize small-batch training. All key hyperparameters, such as dropout ratios and contrastive loss weights, were determined during the first-stage task and kept consistent across all second-stage fine-tuning tasks to ensure the fairness and validity of model transfer.

# Acknowledgements

We sincerely express our gratitude to all members of our research group for their valuable efforts and guidance, as well as to the reviewers for their constructive comments and suggestions.

# Data availability

Codes and datasets are available at https://github.com/wendy1031/AVP-Fusion